\pdfoutput=1
\documentclass[graybox]{svmult}

\usepackage{mathptmx}       
\usepackage{helvet}         
\usepackage{courier}        
\usepackage{type1cm}        

\usepackage{makeidx}         
\usepackage{graphicx}        
\usepackage{multicol}        
\usepackage[bottom]{footmisc}

\usepackage[numbers]{natbib}

\usepackage{multirow}

\usepackage{amsmath}
\usepackage{amsfonts}

\usepackage{tcolorbox}

\usepackage{threeparttable}

\makeindex             

\begin{document}
\newcommand{\KIT}{Karlsruhe Institute of Technology, Institute for Anthropomatics and Robotics, Karlsruhe, Germany}
\newcommand{\FirstAuthor}{Stefan Constantin}
\newcommand{\SecondAuthor}{Jan Niehues}
\newcommand{\ThirdAuthor}{Alex Waibel}
\newcommand{\TitleName}{Multi-task learning to improve natural language understanding}

\newcommand{\AbstractText}{
Recently advancements in sequence-to-sequence neural network architectures have led to an improved natural language understanding.
When building a neural network-based Natural Language Understanding component, one main challenge is to collect enough training data.
The generation of a synthetic dataset is an inexpensive and quick way to collect data.
Since this data often has less variety than real natural language, neural networks often have problems to generalize to unseen utterances during testing.

In this work, we address this challenge by using multi-task learning.
We train out-of-domain real data alongside in-domain synthetic data to improve natural language understanding.

We evaluate this approach in the domain of airline travel information with two synthetic datasets.
As out-of-domain real data, we test two datasets based on the subtitles of movies and series.
By using an attention-based encoder-decoder model, we were able to improve the \(F_1\)-score over strong baselines from 80.76\,\% to 84.98\,\% in the smaller synthetic dataset.
}

\title*{\TitleName}
\titlerunning{Multi-task learning to improve natural language understanding}
\author{\FirstAuthor \and \SecondAuthor \and \ThirdAuthor}
\institute{\FirstAuthor \and \SecondAuthor\and \ThirdAuthor
\at \KIT \\
\email{\texttt{firstname.lastname@kit.edu}}}
%
%
\maketitle

\abstract{\AbstractText}

\abstract*{\AbstracText}


\section{Introduction}
\label{sec:Introduction}
One of the main challenges in building a Natural Language Understanding (NLU) component for a specific task is the necessary human effort to encode the task's specific knowledge.
In traditional NLU components, this was done by creating hand-written rules.
In today's state-of-the-art NLU components, significant amounts of human effort have to be used for collecting the training data.
For example, there are a lot of possibilities to express the situation that someone wants to book a flight from New York to Pittsburgh.
In order to get a good NLU component, we need to have seen many of them in the training data.
Although more and more data has been collected and datasets with this data have been published \cite{SerbanLHCP18}, the datasets often consist of data from another domain, which is needed for a certain NLU component.

An inexpensive and quick way to collect data for a domain is to generate a synthetic dataset where templates are filled with various values.
A problem with such synthetic datasets is to encode enough variety of natural language to be able to generalize to unseen utterances during training.
To do this, an enormous amount of effort will be needed.
In this work, we address this challenge by combining task-specific synthetic data and real data from another domain.
The multi-task framework enables us to combine these two knowledge sources and therefore improve natural language understanding.

In this work, the NLU component is based on an attention-based encoder-decoder model \cite{BahdanauCB15}.
We evaluate the approach on the commonly used travel information task and used as an out-of-domain task the subtitles of movies and series.

\section{Related Work}
There are many appropriate architectures for end-to-end trainable goal-oriented dialog systems \cite{BahdanauCB15, ConstantinNW18, SerbanAHL2017} with different approaches for the NLU part; however, what they have in common is that they need a huge amount of training data.

Multi-task learning has been performed in many machine learning applications, e.\,g., in facial landmark detection an application in the area of vision \cite{ZhangLLT2014}.

Multi-task learning for sequence-to-sequence models in Natural Language Processing is described in \cite{LuongLSVK2016, NiehuesC17, PhamSSHNW2017}.
In \cite{LuongLSVK2016}, machine translation was trained together with either syntax parsing or image captioning on a not attention-based encoder-decoder model.
The encoder was shared between the tasks.
They improved the translation between English and German by up to 1.5 BLEU points.
In \cite{NiehuesC17}, the authors used an attention-based encoder-decoder model and were also able to improve on this model machine translation by up to 1.5 BLEU points by combining machine translation with part-of-speech tagging and named entity recognition (the encoder was shared).
In addition, they presented different architectures for multi-task learning, such as sharing in addition to the encoder, the attention layer, or decoder.
In \cite{PhamSSHNW2017}, the authors used multi-task learning to learn to translate 20 individual languages with one system.

\section{Multi-task Learning}
In the multi-task learning approach of this work, in-domain synthetic data and out-of-domain real data are jointly trained.
In synthetic datasets, there are often missing expressions for situations.
However, in larger out-of-domain datasets, there are expressions for similar situations.
Through the joint training of the encoding for both tasks, we expect a better natural language understanding in the in-domain task because it can be learned to encode situations independent to their expression in natural language. 

\subsection{Architecture}
We use an attention-based encoder-decoder model for multi-task learning.
We share between the tasks the embedding layer and the encoder.
The remaining components of the attention-based encoder-decoder model - the attention layer and the decoder with its final softmax layer - are not shared.
The intuition behind this is, that in our synthetic datasets, there are missing expressions for situations that are in the out-of-domain datasets.
With the training of the out-of-domain datasets, we want to learn to encode situations independent to their expression in natural language.
For improving encoding, we expect the best results by only sharing the encoder because knowledge from the out-of-domain dataset is transfered to the in-domain dataset.

\begin{figure}
    \includegraphics[width=\textwidth]{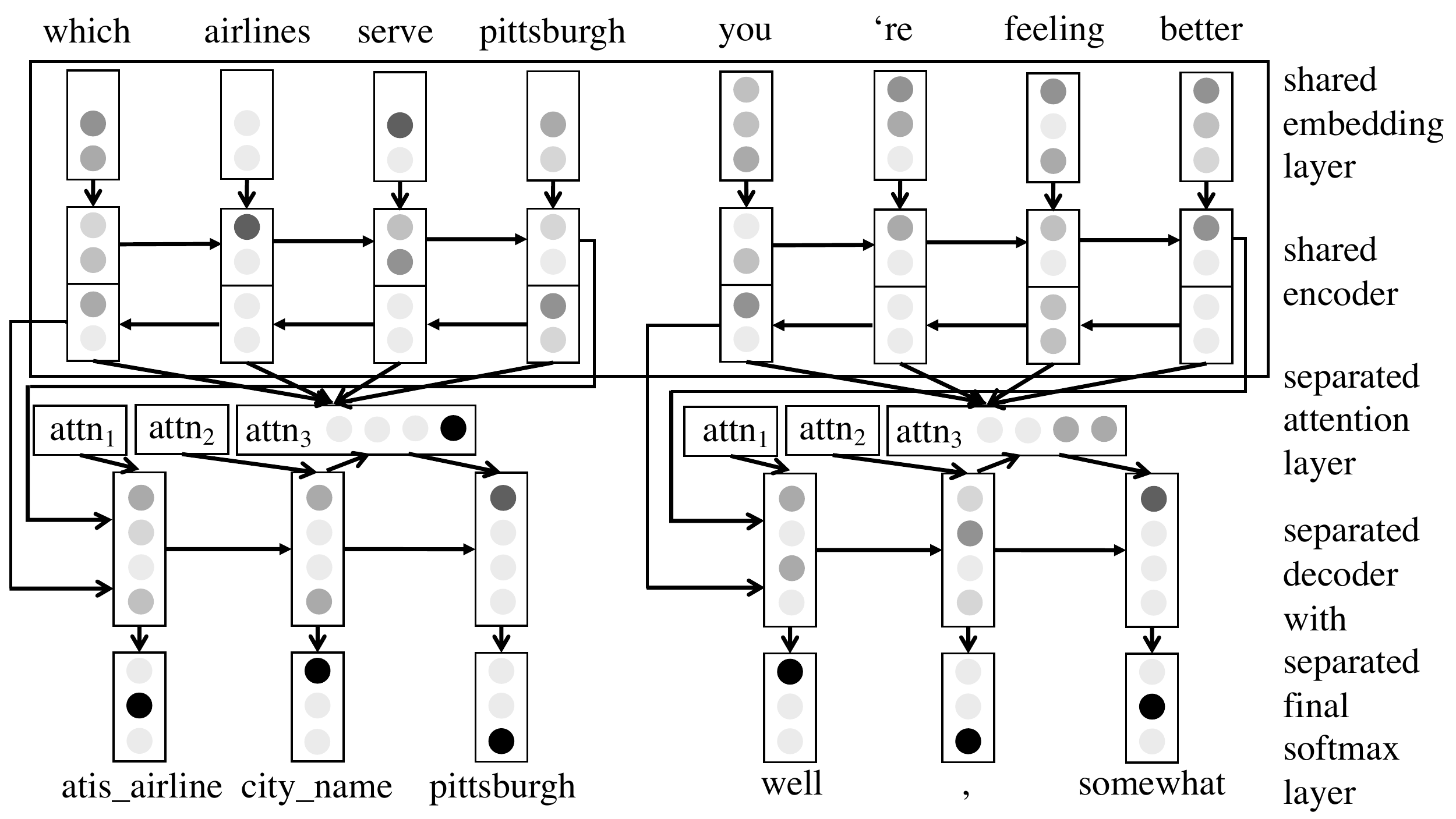}
    \caption{attention-based encoder-decoder}
    \label{fig:EncDec}
\end{figure}

In \cite{PhamSSHNW2017}, an attention-based encoder-decoder model that is able to share the weights of layers between tasks is described and its implementation was published.
We added to this implementation an option to train instances of the smallest dataset \(m\)-times and an option to accumulate gradients and published\footnote{available at \url{https://github.com/isl-mt/OpenNMT-py/tree/MultiTask}} the additions under the MIT license.
The architecture is depicted in Figure \ref{fig:EncDec}.

\subsection{Training Schedule}
In \cite{NiehuesC17}, only one task in each mini-batch is considered because this is more GPU-efficient given that not all weights are shared between the tasks.
Let \(n\) be the number of instances that are trained simultaneously on the GPU.
The instances of one task are grouped into groups of size \(n\).
These groups are randomly shuffled before every epoch during training.
However, in our experiments, updating the weights after the training of a group of one task led to perplexity jumps.
To avoid these jumps, we accumulate the gradients and update our weights only after \(t\) groups.
This means that our mini-batch size is \(t \cdot n\).
We use the Adam optimization algorithm \cite{KingmaB2015} for updating the weights.

After the multi-task learning, we fine-tune the model by retraining the model only with the synthetic dataset.
For this fine-tuning, we reset all the parameters of the Adam optimization algorithm.

The out-of-domain datasets have a huge size in comparison to the synthetic datasets.
To avoid instances of the synthetic datasets are not considered in the training of the model, instances of the synthetic dataset are trained \(m\)-times during one epoch.

\section{Experimental Setup}

\subsection{Data}
For the out-of-domain task, we use two subsets of the English OpenSubtitle corpus \cite{Tiedemann2009}\footnote{based on http://www.opensubtitles.org/} in this work.
The OpenSubtitle corpus consists of the subtitles of movies and series.
The first subset was published by \cite{Senellart2017}\footnote{available at \url{https://s3.amazonaws.com/opennmt-trainingdata/opensub\_qa\_en.tgz}} and consists of all the sentence pairs from the OpenSubtitle corpus that have the following properties: the first sentence ends with a question mark; the second sentence follows directly the first sentence and has no question mark; and the time difference between the sentences is less than 20 seconds.
In total, the subset has more than 14 million sentence pairs for training and 10\,000 sentence pairs for validation.
In the following sections, this dataset is called \textit{OpenSubtitles QA}.
We created the second subset in a similar manner as the SubTle dataset \cite{AmeixaC2013} was created.
It consists of sentence pairs with the following properties: the second sentence follows directly the first sentence; both sentences end with a point, exclamation point, or question mark; and between the two sentences, there is at maximum a pause of 1 second.
In the following sections, this dataset is called \textit{OpenSubtitles dialog}.
To be able to train the attention-based encoder-decoder model in a reasonable time, we only used the first 14 million sentence pairs for training.
The next 10\,000 sentence pairs were used for validation.
For both datasets we used the default English word tokenizer of the Natural Language Toolkit (NLTK) \cite{BirdKL2009}\footnote{https://www.nltk.org/} for tokenization.
As there is another tokenization approach in the OpenSubtitle corpus in comparison to the tokenizer in the NLTK, we had to merge the tokens 's, 're, 't, 'll, and 've to their previous token in the \textit{OpenSubtitles dialog} dataset to improve the compatibility with the tokenization of the NLTK.

We generated two synthetic datasets.
These two datasets are based on a subset of the ATIS (Airline Travel Information Systems) dataset \cite{Price1990} that was published by \cite{HakkaniTCCGDW2016}\footnote{available at https://github.com/yvchen/JointSLU} and is called \textit{ATIS real} in the following sections.
In the ATIS corpus, every user utterance has one or multiple intents and every word of a user utterance is tagged in the IOB format.
The format is depicted in Figure \ref{fig:ATISS2S}.
However, the out-of-domain dataset is no intent and slot filling task.
It is a sequence-to-sequence task.
To train both tasks together, we converted the intent and slot filling task to a sequence-to-sequence task.
The target sequence consists of the intents followed by the parameters.
A parameter consists of the slot name and the slot value.
An example conversion is depicted in Figure \ref{fig:ATISS2S}.
With this conversion, the attention-based encoder-decoder can learn both tasks: user utterance to semantic representation and subtitle sentence A to subtitle sentence B.
In Figure \ref{fig:EncDec}, it is depicted how the attention-based encoder-decoder handles both tasks.

\begin{figure}

    \begin{tabular}{ | p{2.15cm} | c | c | c | c | c | c | c | c | c | }
    	\hline
    	utterance (source sequence) & show & me & flights & between & new & york & city & and & pittsburgh \\
    	\hline
    	slots & O & O & O & O & B-fromloc & I-fromloc & I-fromloc & O & B-toloc  \\
    	\hline
    	intents & \multicolumn{9}{c |}{ATIS\_flight} \\
    	\hline
    	target sequence & \multicolumn{9}{c |}{ATIS\_flight fromloc new york city toloc pittsburgh} \\
    	\hline
    \end{tabular}
    
    \caption{format of the ATIS corpus and the conversion to a sequence-to-sequence problem}
    \label{fig:ATISS2S}
\end{figure}

In the \textit{ATIS real} dataset, there are 4478 tagged user utterances for training, 500 for validation and 893 for testing.

For training, the smaller synthetic dataset has 212 templates that form 17\,679 source target sequence pairs after filling the template placeholders and is called \textit{ATIS small} in the following sections and the larger dataset has 832 templates that form 70\,040 source target sequence pairs and is called \textit{ATIS medium} in the following sections.
The validation and test utterances are the same in all three datasets (\textit{ATIS real}, \textit{ATIS small}, and \textit{ATIS medium}).
The templates for the training utterances of the \textit{ATIS small} dataset were generated by extracting all the sequences that have a new parameter in the target sequence that was not included in any target sequence extracted before.
Extracting all the sequences that have a parameter combination that was not included in any target sequence extracted before, form the training templates of the \textit{ATIS medium} dataset.
In the extracted sequences, the parameter values were replaced by placeholders to become templates.
For the placeholders, all the possible values were inserted.
When one template produced more than 1000 source target sequence pairs, then, instead of the Cartesian product, the random permutation algorithm \cite{Hazwani2016} was used, which produces as many source target sequence pairs as the values of the placeholder with the greatest number of values.
For both datasets, we alphabetically sorted the parameters to ease the learning process.

\subsection{Evaluation}
We evaluate the quality of the predicted intents and parameters with the metric \(F_1\)-score.
Every intent and parameter is considered individually.
For averaging the \(F_1\)-score over the target sequences, we use micro-averaging.
This means that we count the number of true positives, false positives, and false negatives for all the intents and parameters and calculate the recall and precision for the \(F_1\)-score with these.

In addition, we provide the metric intent accuracy.
For the intent accuracy, the number of completely correct predicted intents (the intents of the reference and hypothesis must be the same) is divided by the number of target sequences.

\subsection{System Setup}
We optimized our single-task baseline to get a strong baseline in order to exclude better results in multi-task learning in comparison to single-task learning only because of these two following points: network parameters suit the multi-task learning approach better and a better randomness while training in the multi-task learning.
To exclude the first point, we tested different hyperparameters for the single-task baseline.
We tested all the combinations of the following hyperparameter values: 256, 512, or 1024 as the sizes for the hidden states of the LSTMs, 256, 512, or 1024 as word embedding sizes, and a dropout of 30\,\%, 40\,\%, or 50\,\%.
We used subword units generated by byte-pair encoding (BPE) \cite{SennrichHB2016} as inputs for our model.
To avoid bad subword generation for the synthetic datasets, in addition to the training dataset, we considered  the validation and test dataset for the generating of the BPE merge operations list.
We trained the configurations for 14 epochs and trained every configuration three times.
We chose the training with the best quality with regard to the validation \(F_1\)-score to exclude disadvantages of a bad randomness.
We got the best quality with regard to the \(F_1\)-score with 256 as the size of the hidden states of the LSTMs, 1024 as word embedding size, and a dropout of 30\,\%.
For the batch size, we used 64.

We optimized our single-task model trained on real data in the same manner as the single-task baseline, except that we used 64 epochs.

In the multi-task learning approach, we trained both tasks for 10 epochs.
We use for \(m\) (the instance multiplicator of the synthetic dataset) such a value that the synthetic dataset has nearly the size of one-tenth of the out-of-domain dataset.
Because of long training times, we were not able to optimize the hyperparameters.
We chose 256 as the size of the hidden states of the LSTMs, 1024 as word embedding size, and 50\,\% for the dropout and were not able to run multiple runs.
For \(n\) (the number of instances that are trained simultaneously on the GPU), we chose 128 and for \(t\) (number of groups after that the model weights are updated) we chose 11.
Other hyperparameters in the single-task and multi-task experiments were not changed from the default values of the published implementation.

We used the best epoch with regard to the validation \(F_1\)-score to fine-tune our model.
To exclude only better results because of good random initialization, we made three runs, used the epoch with the best validation \(F_1\)-score from every run, and chose the run with the worst validation \(F_1\)-score for evaluation.
We used 64 as the batch size, 50\,\% as dropout, and 14 as the number of epochs.

We used subword units generated by BPE for all approaches and used 40\,000 as the limit for the number of BPE merging operations as well as the vocabulary size.

\section{Results}
\label{sec:Results}
For all validation and test results, the validation and test dataset of the \textit{ATIS real} dataset is used.

\begin{figure}
    \includegraphics{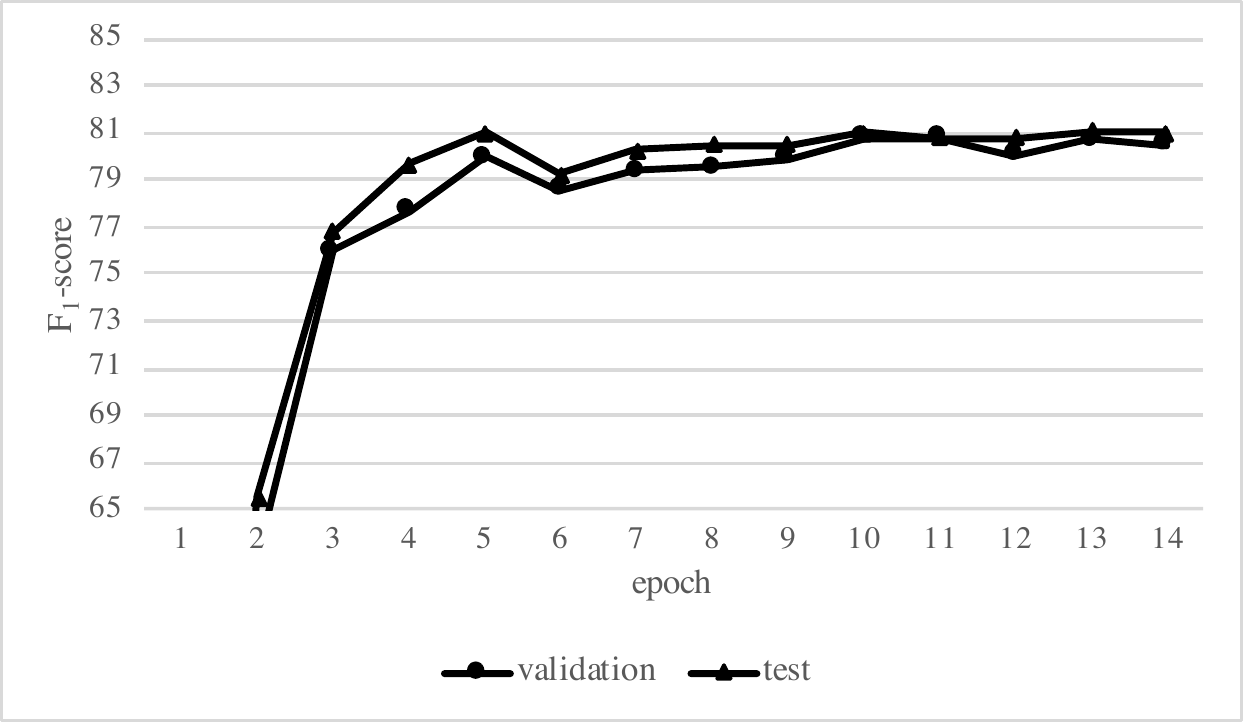}
    \caption{validation and test \(F_1\)-score of the single-task baseline trained with the \textit{ATIS small} dataset}
    \label{fig:singleTaskSmallATIS}
\end{figure}

\begin{figure}
    \includegraphics{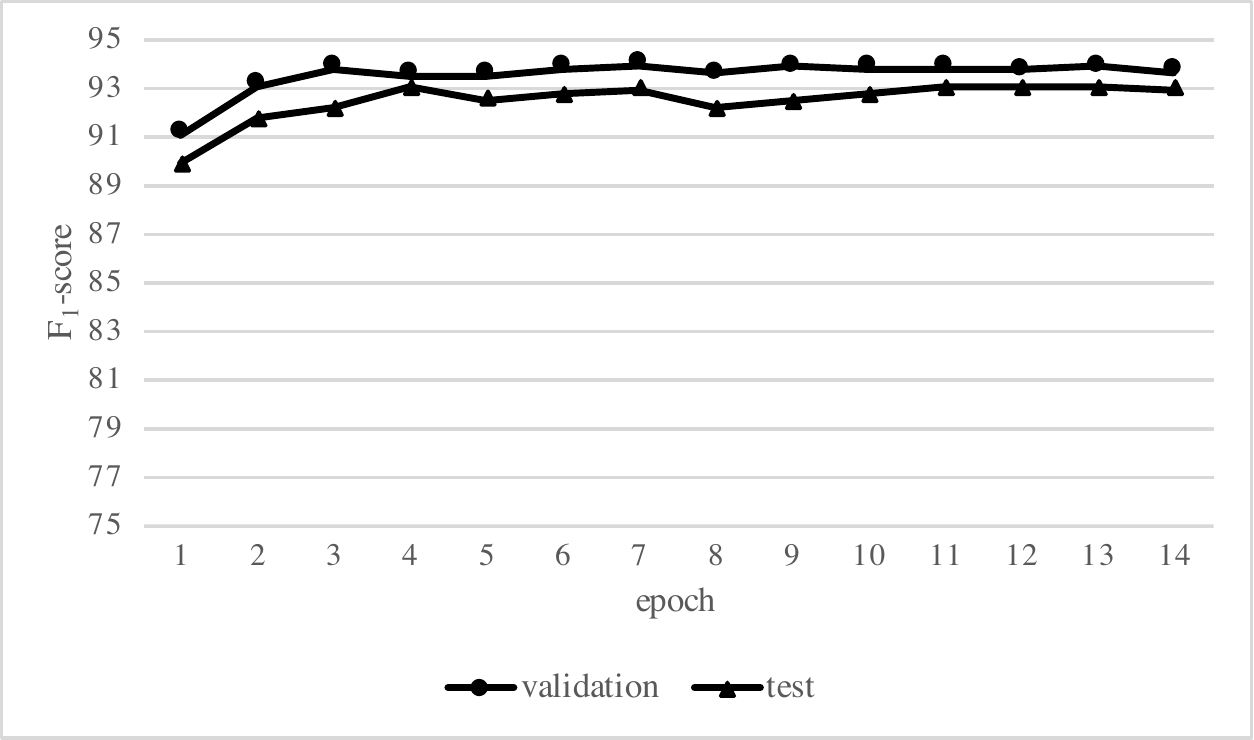}
    \caption{validation and test \(F_1\)-score of the single-task baseline trained with the \textit{ATIS medium} dataset}
    \label{fig:singleTaskATIS}
\end{figure}

In Figure \ref{fig:singleTaskSmallATIS}, the test \(F_1\)-score of the training run of the configuration with the best validation \(F_1\)-score is depicted with respect to the epoch for the \textit{ATIS small} dataset and in Figure \ref{fig:singleTaskATIS} for the \textit{ATIS medium} dataset.
The best result is achieved after epoch 11 or 7, respectively.
There is no trend for a further improvement after epoch 14.
The test \(F_1\)-score of the best epoch according to the validation \(F_1\)-score is depicted in the Tables \ref{tab:ResultsATISSmall} and \ref{tab:ResultsATIS}, respectively.

In Table \ref{tab:ResultsATISSmall}, the validation and test \(F_1\)-scores and intent accuracies with regard to the best validation \(F_1\)-score of the multi-task learning approach with the \textit{ATIS small} dataset is depicted.
The test \(F_1\)-score could be improved 2.32 percentage points with not fine-tuned multi-task learning with the \textit{OpenSubtitles QA} dataset and 4.22 percentage points to 84.98\,\% with the \textit{OpenSubtitles dialog} dataset.
The test intent accuracies could be improved with not fine-tuned multi-task learning 5.60 and 4.93 percentage points, respectively.
For both out-of-domain datasets, fine-tuning did change the \(F_1\)-score only neglectable.

In Table \ref{tab:ResultsATIS}, the validation and test \(F_1\)-scores and intent accuracies with regard to the best validation \(F_1\)-score of the multi-task learning approach with the \textit{ATIS medium} dataset is depicted.
The test \(F_1\)-score could be improved 0.52 percentage points with not fine-tuned multi-task learning with the \textit{OpenSubtitles QA} dataset and 0.30 percentage points with the \textit{OpenSubtitles dialog} dataset.
The test intent accuracies could be improved with not fine-tuned multi-task learning by 0.34 and 1.79 percentage points, respectively.
These improvements are not big, but the \(F_1\)-score of the multi-task learning with the \textit{OpenSubtitles QA} dataset is only 0.13 percentage points below the results of the model trained on the complete real training data of the \textit{ATIS real} dataset.

\begin{table}
	\centering

	\begin{tabular}{ l | l | c c | c c }
	\hline
	training dataset(s) & model & \multicolumn{2}{c |}{validation (\textit{ATIS real})} & \multicolumn{2}{c}{test (\textit{ATIS real})} \\
	\cline{3-6}
	&  & \(F_1\) & intent acc & \(F_1\) & intent acc \\
	\hline
	\textit{ATIS small} & single-task baseline & 80.79 & 86.00 & 80.76 & 82.64 \\	
	\hline
	\multirow{2}{*}{\textit{ATIS small} + \textit{OpenSubtitles QA}} & shared encoder & 82.21 & \textbf{87.60} & 83.08 & 88.24 \\
	& shared encoder fine-tuned & 82.46 & 87.00 & 83.06 & 87.68 \\
	\hline
	\multirow{2}{*}{\textit{ATIS small} + \textit{OpenSubtitles dialog}} & shared encoder & 82.11 & 86.00 & \textbf{84.98} & 87.57 \\
	& shared encoder fine-tuned & \textbf{82.65} & 83.80 & 84.55 & \textbf{88.80} \\
	\hline
	\end{tabular}

	\caption{results on the \textit{ATIS real} dataset of the systems trained with the \textit{ATIS small} dataset}
	\label{tab:ResultsATISSmall}
\end{table}

\begin{table}
	\centering

	\begin{tabular}{ l | p{2.4cm} | c c | c c }
	\hline
	training dataset(s) & model & \multicolumn{2}{c |}{validation (\textit{ATIS real})} & \multicolumn{2}{c}{test (\textit{ATIS real})} \\
	\cline{3-6}
	&  & \(F_1\) & intent acc & \(F_1\) & intent acc \\
	\hline
	\textit{ATIS medium} & single-task baseline & 93.96 & 95.40 & 92.97 & 94.96 \\
	\hline
	\multirow{2}{*}{\textit{ATIS medium} + \textit{OpenSubtitles QA}} & shared encoder & 93.80 & 96.40 & \textbf{93.49} & 95.30 \\
	& shared encoder fine-tuned & \textbf{94.00} & \textbf{97.20} & 92.81 & 94.96 \\
	\hline
	\multirow{2}{*}{\textit{ATIS medium} + \textit{OpenSubtitles dialog}} & shared encoder & 93.74 & 96.40 & 93.27 & \textbf{96.75} \\
	& shared encoder fine-tuned & 93.88 & 97.00 & 92.88 & 96.42 \\
	\hline
	\textit{ATIS real} & single-task trained on real data & \textbf{95.97} & 96.80 & \textbf{93.62} & 94.74 \\
	\hline
	\end{tabular}

	\caption{results on the \textit{ATIS real} dataset of the systems trained with the \textit{ATIS medium} dataset}
	\label{tab:ResultsATIS}
\end{table}

\section{Conclusions and Further Work}
\label{sec:Conlusion}
In this work, we evaluated whether the training of a synthetic dataset alongside with an out-of-domain dataset can improve the quality in comparison to train only with the synthetic dataset.
Although we optimized the model of the single-task learning baseline and not the model of the multi-task learning approach, we were able to increase the \(F_1\)-score 4.22 percentage points to 84.98\,\% for the smaller synthetic dataset (\textit{ATIS small}).
For the bigger dataset (\textit{ATIS medium}), we could not significantly improve the results, but the results are already in the near of the results of the model trained on the real data.
To improve the quality of dialog systems for these exist only strong under-resourced synthetic datasets is especially helpful because the better a system is, the more it encourages users to use it.
This is often an inexpensive way to collect data to log real user usage.
However, by collecting real user data, it is necessary to account privacy laws.

The problem with the \textit{OpenSubtitles QA} dataset is, that the form question as source sequence and answer as target sequence differs from the form of the ATIS datasets.
The problem with the \textit{OpenSubtitles dialog} dataset is that it is very noisy.
Responses do not often refer to the previous utterance.
In future work, it would be interesting to test other datasets or a combination of datasets whose form is better fitting or are less noisy, respectively.

We expect a further improvement of the multi-task learning approach by optimizing the parameters of our model in the multi-task learning approach.
However, this is very computation time intensive because the out-of-domain datasets have 14 million instances, and therefore, we leave it open for future work.

We evaluated the multi-task learning approach with the attention-based encoder-decoder model, but we also expect an improvement by the multi-task learning approach for other architectures, such as the transformer model \cite{VaswaniSPUJGKP2017}, which could be researched in future work.


\section*{Acknowledgement}
This work has been conducted in the SecondHands project which has received funding from the European Union’s Horizon 2020 Research and Innovation programme (call:H2020- ICT-2014-1, RIA) under grant agreement No 643950.
\bibliographystyle{spmpsci.bst}
\bibliography{bibliography}

\end{document}